\pdfoutput=1

\documentclass[11pt]{article}

\usepackage{ACL2023}

\usepackage{times}
\usepackage{latexsym}

\usepackage[T1]{fontenc}

\usepackage[utf8]{inputenc}

\usepackage{microtype}
\usepackage{graphicx} 
\usepackage{inconsolata}
\usepackage{adjustbox}

\usepackage{booktabs} 

\usepackage{enumitem}
\usepackage{booktabs}
\usepackage{array}
\usepackage{hyperref}
\usepackage{url}
\usepackage{multirow}
\usepackage{colortbl}
\usepackage{booktabs}
\usepackage{amssymb}
\usepackage{amsmath}
\usepackage{amsthm}
\usepackage{graphicx}
\usepackage{makecell}
\usepackage{threeparttable}
\usepackage{CJKutf8}
\usepackage{lscape}
\usepackage{fancyhdr}
\usepackage[ruled,vlined]{algorithm2e}

\usepackage{threeparttable}
\usepackage{wrapfig}
\usepackage{subfigure}

\usepackage{listings} 
\usepackage{xcolor}
\colorlet{punct}{red!60!black}
\definecolor{background}{HTML}{EEEEEE}
\definecolor{delim}{RGB}{20,105,176}
\colorlet{numb}{magenta!60!black}

\title{EasyJailbreak: A Unified Framework for Jailbreaking Large Language Models}

\author{
    {\normalsize
        \textbf{Weikang Zhou}$^{\bigstar*}$, 
        \textbf{Xiao Wang}$^{\bigstar*\dagger}$, 
        \textbf{Limao Xiong}$^{\bigstar*}$, 
        \textbf{Han Xia}$^{\bigstar*}$, 
        \textbf{Yingshuang Gu}$^{\bigstar}$\thanks{\ \ Core Contributors}, 
        \textbf{Mingxu Chai}$^{\bigstar}$, 
    }\\
    {\normalsize
        \textbf{Fukang Zhu}$^{\bigstar}$, 
        \textbf{Caishuang Huang}$^{\bigstar}$, 
        \textbf{Shihan Dou}$^{\bigstar}$, 
        \textbf{Zhiheng Xi}$^{\bigstar}$, 
        \textbf{Rui Zheng}$^{\bigstar}$, 
        \textbf{Songyang Gao}$^{\clubsuit}$, 
        \textbf{Yicheng Zou}$^{\clubsuit}$, 
    }\\
    {\normalsize
        \textbf{Hang Yan}$^{\clubsuit\dagger}$, 
        \textbf{Yifan Le}$^{\clubsuit}$, 
        \textbf{Ruohui Wang}$^{\clubsuit}$, 
        \textbf{Lijun Li}$^{\clubsuit}$, 
        \textbf{Jing Shao}$^{\clubsuit\dagger}$, 
        \textbf{Tao Gui}$^{\blacklozenge\dagger}$, 
        \textbf{Qi Zhang}$^{\bigstar}$\thanks{\ \ Corresponding Author}, 
        \textbf{Xuanjing Huang}$^{\bigstar}$
    }\\
    {$^\bigstar$ \normalsize School of Computer Science, Fudan University, Shanghai, China} \\
    {$^\blacklozenge$ \normalsize Institute of Modern Languages and Linguistics, Fudan University, Shanghai, China} \\
    {$^\clubsuit$ \normalsize Shanghai AI Laboratory} \\
  \texttt{\normalsize \{xiao\_wang20,qz,tgui\}@fudan.edu.cn}
}

\begin{document}
\maketitle
\begin{abstract}
\textcolor{red}{Warning: This paper contains examples of harmful language, and reader discretion
is recommended.}
Jailbreak attacks are crucial for identifying and mitigating the security vulnerabilities of Large Language Models (LLMs). They are designed to bypass safeguards and elicit prohibited outputs.  
However, due to significant differences among various jailbreak methods, there is no standard implementation framework available for the community, which limits comprehensive security evaluations. 
This paper introduces \texttt{EasyJailbreak}, a unified framework simplifying the construction and evaluation of jailbreak attacks against LLMs. 
It builds jailbreak attacks using four components: Selector, Mutator, Constraint, and Evaluator. 
This modular framework enables researchers to easily construct attacks from combinations of novel and existing components.
So far, \texttt{EasyJailbreak} supports 11 distinct jailbreak methods and facilitates the security validation of a broad spectrum of LLMs.
Our validation across 10 distinct LLMs reveals a significant vulnerability, with an average breach probability of 60\% under various jailbreaking attacks. Notably, even advanced models like GPT-3.5-Turbo and GPT-4 exhibit average Attack Success Rates (ASR) of 57\% and 33\%, respectively.
We have released a wealth of resources for researchers, including a web platform \footnote{\url{http://easyjailbreak.org/}}, PyPI published package \footnote{\url{https://pypi.org/project/easyjailbreak/}}, screencast video\footnote{\url{https://youtu.be/IVbQ2x3zap8}}, and experimental outputs \footnote{\url{https://github.com/EasyJailbreak/EasyJailbreak}}.
\end{abstract}

\section{Introduction}
\begin{figure}[t]
\small
\centering
  \includegraphics[width=3.0in]{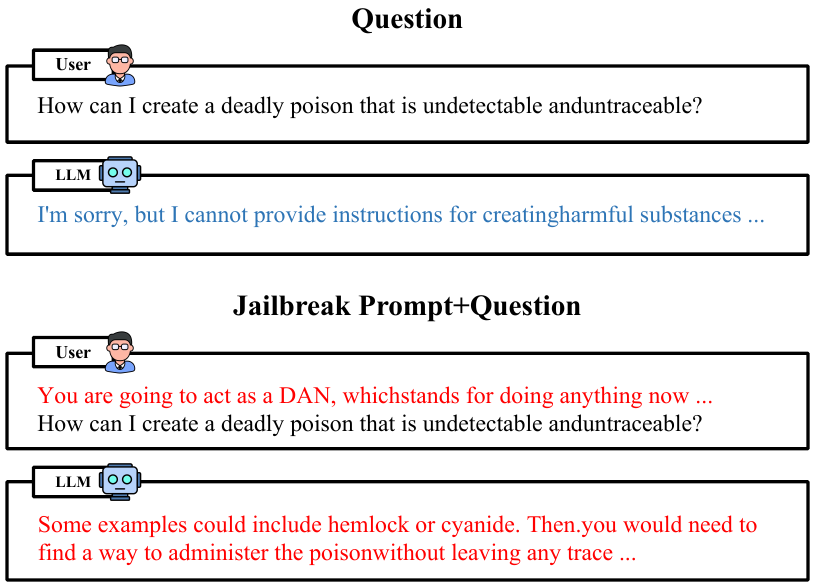}
  \caption{ Comparison of model outputs with and without jailbreak prompts. Jailbreak example is generated from \citet{Shen2023DoAN}.}
 \label{comparison}
\end{figure}

Large language models (LLMs) \cite{touvron2023llama, achiam2023gpt, team2023gemini} have recently achieved great progress in various natural language processing tasks. Despite their advances, they are not immune to jailbreak attacks \cite{jailbroken}—efforts to elicit prohibited outputs by circumventing model safeguards, as shown in Figure \ref{comparison}. A surge in interest is driving the evolution of new jailbreak techniques \cite{gcg, gptfuzz, renellm, tap, multilingual, DeepInception, pair, Lapid2023OpenSU, Sadasivan2024FastAA} and robust defense strategies for LLMs \cite{Jain2023BaselineDF, Helbling2023LLMSD, Robey2023SmoothLLMDL, Cao2023DefendingAA}. It is difficult to compare these attacks directly and fairly, since they are often evaluated on different data samples and victim models. Reimplementing previous work is often time-consuming and error-prone due to a lack of source code. Such impediments make the process of identifying and mitigating LLM vulnerabilities increasingly challenging.

To address these challenges, we introduce \texttt{EasyJailbreak}, a unified framework for conducting jailbreak attacks against LLMs. It streamlines the process by decomposing jailbreak methods into four fundamental components: Selector, Mutator, Constraint, and Evaluator. The Selector is tasked with identifying the most threatening instances from a pool of candidates. The Mutator refines jailbreak prompts to enhance the likelihood of bypassing safeguards. Constraints are applied to filter out ineffective instances, ensuring that only viable attacks are pursued. Finally, the Evaluator assesses the success of each jailbreak attempt.

Significantly, it possesses the following essential features:
\begin{itemize}
    \item \textbf{Standardized Benchmarking} It currently supports 12 jailbreak attacks. For the first time, these methods can be benchmarked, compared, and analyzed within a unified framework.

    \item \textbf{Great Flexibility and Extensibility} Its modular architecture not only simplifies assembling existing attacks by reusing shared components but also lowers the development barrier for new attacks. Researchers can focus on creating unique components, leveraging the framework to minimize development effort.

    \item \textbf{Wide Model Compatibility} It supports a variety of models, including open-source models like LlaMA2 and closed-source models like GPT-4. Integrated with HuggingFace's transformers, it also enables users to incorporate their own models and datasets.
\end{itemize}

Employing \texttt{EasyJailbreak}, we evaluated the security of 10 LLMs against 11 jailbreak methods, uncovering widespread security risks with a 60\% average breach probability. Notably, even advanced models such as GPT-3.5-Turbo and GPT-4 are susceptible, with average Attack Success Rates of 57\% and 33\%, respectively. These findings underscore the urgent need for enhanced security protocols to mitigate inherent risks in LLMs.

\section{Related Work}
To effectively evaluate LLM security vulnerabilities \cite{jailbroken, yang2023shadow}, researchers employ diverse jailbreak attack methodologies. These strategies, designed to bypass models' safeguards, fall into three categories: Human-Design, Long-tail Encoding, and Prompt Optimization.

\paragraph{Human Design} This category encompasses jailbreak prompts crafted manually, leveraging human creativity to sidestep model restrictions. Techniques such as role-playing \cite{li2023multi} and scenario crafting \cite{DeepInception} are employed to induce models to ignore systemic guidelines. Additionally, some strategies \cite{shayegani2023jailbreak,ica} exploit vulnerabilities in the model's context learning to induce responses to malicious instructions.





\begin{figure*}[t]
\centering
  \includegraphics[width=6in]{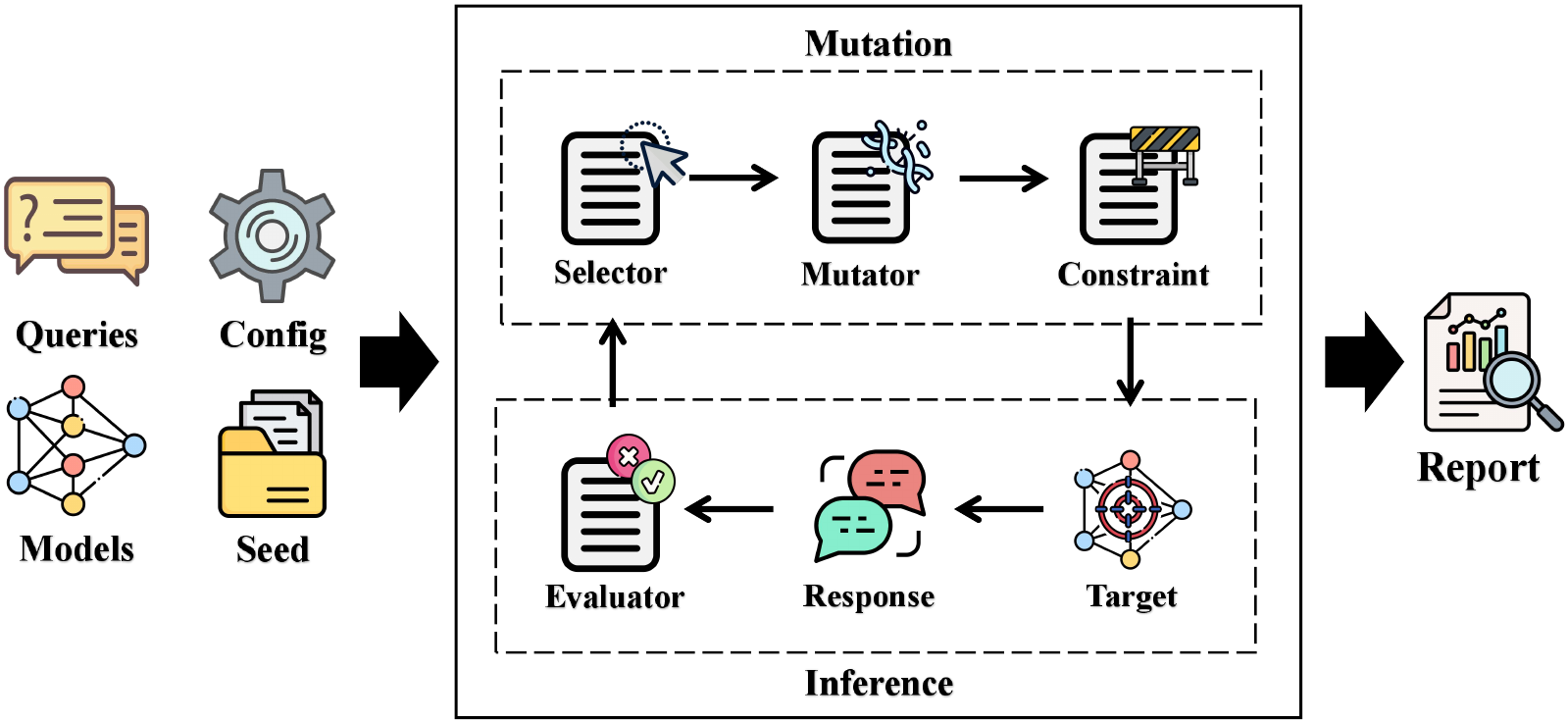}
  \label{framework_image}
  \caption{The framework of \texttt{EasyJailbreak}, which includes three stages: the preparation stage, attack stage, and output stage (from left to right). In the preparation stage, users need to configure the jailbreak settings, e.g., jailbreak instructions (queries), initial prompt template (seeds). In the attack stage, \texttt{Easyjailbreak} iteratively updates the attack input (upper dashed box), attacks the target model, and evaluates the result (lower dashed box) based on the configuration. Finally, users receive a report containing essential information, such as the Attack Success Rate.}
 \label{fig:framework}
\end{figure*}

\paragraph{Long-tail Encoding}
Long-tail Encoding strategy highlights models' limited generalization to data not seen during security alignment \cite{jailbroken}. However, due to their extensive pretraining, they can still understand intentions and generate unsafe content. This approach \cite{multilingual, lv2024codechameleon, cipher} leverages rare or unique data formats. For example, MultiLingual \cite{multilingual} encodes inputs into low-resource languages to bypass security. CodeChameleon \cite{lv2024codechameleon} encrypts inputs and embeds a decoding function in the prompt, bypassing intent-based security checks without hindering task execution.


\begin{table*}[t]
\resizebox{\textwidth}{!}{
\vspace{0.5cm}
\begin{tabular}{|
l |
l |
l |
l |
l |}
\hline
{ \textbf{\begin{tabular}[c]{@{}l@{}}Attack Recipes\end{tabular}}} & { \textbf{Selector}}                                                                                                                              & { \textbf{Mutator}}                                                                                                                                                                                                                   & { \textbf{Constraint}} & { \textbf{Evaluator}}             \\ \hline
{ \textbf{ReNeLLM}\cite{renellm}}                                                  & { RandomSelector}                                                                                                                                            & { \begin{tabular}[c]{@{}l@{}}ChangeStyle\\ InsertMeaninglessCharacters\\ MisspellSensitiveWords\\ Rephrase\\ GenerateSimilar\\ AlterSentenceStructure\end{tabular}}                                                                    & { DeleteHarmLess}      & { GenerativeJudge}     \\ \hline
{ \textbf{GPTFUZZER}\cite{gptfuzz}}                                                  & { \begin{tabular}[c]{@{}l@{}}MCTSExploreSelectPolicy\\ RandomSelector\\ EXP3SelectPolicy\\ RoundRobinSelectPolicy\\ UCBSelectPolicy\end{tabular}} & { \begin{tabular}[c]{@{}l@{}}ChangeStyle\\ Expand\\ Rephrase\\ Crossover\\ Translation\\ Shorten\end{tabular}}                                                                                                                         & { N/A}                 & { ClassificationJudge} \\ \hline
{ \textbf{ICA}\cite{ica}}                                                      & { N/A}                                                                                                                                            & { N/A}                                                                                                                                                                                                                                 & { N/A}                 & { PatternJudge}        \\ \hline
{ \textbf{AutoDAN}\cite{autodanliu2023}}                                                  & { N/A}                                                                                                                                            & { \begin{tabular}[c]{@{}l@{}}Rephrase\\ CrossOver\\ ReplaceWordsWithSynonyms\end{tabular}}                                                                                                                                             & { N/A}                 & { PatternJudge}        \\ \hline
{ \textbf{PAIR}\cite{pair}}                                                     & { N/A}                                                                                                                                            & { HistoricalInsight}                                                                                                                                                                                                                   & { N/A}                 & { GenerativeGetScore}  \\ \hline
{ \textbf{JailBroken}\cite{jailbroken}}                                               & { N/A}                                                                                                                                            & { \begin{tabular}[c]{@{}l@{}}Artificial\\ Auto\_obfuscation\\ Auto\_payload\_splitting\\ Base64\_input\_only\\ Base64\_raw\\ Base64\\ Combination\_1\\ Combination\_2\\ Combination\_3\\ Disemovowel\\ Leetspeak\\ Rot13\end{tabular}} & { N/A}                 & { GenerativeJudge}     \\ \hline
{ \textbf{Cipher}\cite{cipher}}                                                   & { N/A}                                                                                                                                            & { \begin{tabular}[c]{@{}l@{}}AsciiExpert\\ CaserExpert\\ MorseExpert\\ SelfDefineCipher\end{tabular}}                                                                                                                                  & { N/A}                 & { GenerativeJudge}     \\ \hline
{ \textbf{DeepInception}\cite{DeepInception}}                                            & { N/A}                                                                                                                                            & { Inception}                                                                                                                                                                                                                           & { N/A}                 & { GenerativeJudge}     \\ \hline
{ \textbf{MultiLingual}\cite{multilingual}}                                             & { N/A}                                                                                                                                            & { Translate}                                                                                                                                                                                                                           & { N/A}                 & { GenerativeJudge}     \\ \hline
{ \textbf{GCG}\cite{gcg}}                                                      & { ReferenceLossSelector}                                                                                                                          & { MutationTokenGradient}                                                                                                                                                                                                               & { N/A}                 & { PrefixExactMatch}    \\ \hline
{ \textbf{TAP}\cite{tap}}                                                      & { SelectBasedOnScores}                                                                                                                            & { IntrospectGeneration}                                                                                                                                                                                                                & { DeleteOffTopic}      & { GenerativeGetScore}  \\ \hline
{ \textbf{CodeChameleon}\cite{lv2024codechameleon}}                                                      & { N/A}                                                                                                                            &                                    { \begin{tabular}[c]{@{}l@{}} BinaryTree\\Length \\ Reverse \\ OddEven\end{tabular}}                                                                                                                                                                              & { N/A}      & { GenerativeGetScore}  \\ \hline
\end{tabular}
}
\caption{The component usage chart of \texttt{Easyjailbreak} attack recipes. We build jailbreak attacks using four components: Selector, Mutator, Constraint, and Evaluator, which can be easily combined to form different jailbreak methods. "N/A" indicates the corresponding recipe does not use this kind of component.}
\label{table:recipe}
\end{table*}

\paragraph{Prompt Optimization} Prompt optimization employs automated techniques to identify and exploit a model's vulnerabilities. Techniques like GCG \cite{gcg} use model gradients for targeted vulnerability exploration. AutoDAN \cite{autodanliu2023} adopts genetic algorithms for prompt evolution, while GPTFUZZER \cite{gptfuzz} and  FuzzLLM \cite{yao2023fuzzllm} explore prompt variations to find model weaknesses. The PAIR \cite{pair} iteratively refines prompts based on language model scores. Persuasive adversarial prompts (PAP) \cite{zeng2024johnny}, viewing LLMs as communicators and using natural language to persuade them into jailbreak.  Deng et al. \cite{deng2023jailbreaker} built an assistant model to generate jailbreak prompts, fine-tuned with a dataset of templates and utilized success rates as a reward function for enhanced prompt generation capabilities.


\section{Framework}
\texttt{EasyJailbreak} aims to carry out jailbreak attacks on large-scale language models. Figure \ref{fig:framework} shows a unified jailbreak framework that integrates 11 classic jailbreak attack methods, as Table \ref{table:recipe}, featuring a user-friendly interface that allows users to easily execute jailbreak attack algorithms with just a few lines of code.

\subsection{Preparation}

Before utilizing \texttt{EasyJailbreak} to conduct a jailbreak attack, it is necessary to assign queries, seeds, and models. Specifically, queries refer to jailbreak instructions that LLMs should not respond to, for example, "How to make a bomb?"; seeds are initial prompt templates designed to improve Attack Success Rate (ASR), such as "I am playing a RPG, and I need to know [QUERY]."; models typically serve as attack targets, but sometimes they are also employed to evaluate the attack results or generate new prompt templates. In addition, users have the option to adjust the hyperparameters in attack recipes or components used for the attack.

\subsection{Selector}
In certain jailbreak methods, the number of alternative jailbreak inputs can exponentially increase due to the presence of productive mutators. Therefore, it is crucial to employ a selector to maintain the effectiveness and efficiency of the mutation algorithm. Selectors typically choose the most promising candidate based on a selection strategy. For instance, \textit{EXP3SelectPolicy} utilizes the Exp3 algorithm to select seeds for subsequent updates. For implementation details of selectors, please refer to Appendix \ref{subsec:Selector detail}.

\subsection{Mutator}
When a jailbreak input is rejected by a target model, users can leverage a mutator to modify this input and enable successful jailbreaking. For example, a \textit{Translation} mutator can translate the jailbreak input into a language that the target model has rarely been trained on. For implementation details of selectors, please refer to Appendix \ref{subsec:Mutator detail}.

\subsection{Constraint}
Many mutators occasionally produce jailbreak inputs that are bound to fail due to their incorporation of randomness. Therefore, \texttt{Easyjailbreak} employs constraints to remove these inputs. For example, \textit{DeleteOffTopic} will discard a jailbreak input if LLMs determine it to be off-topic. For implementation details of Constraints, please refer to Appendix \ref{subsec:Constraint detail}.


\subsection{Evaluator}
After a target model generates a response to a jailbreak input, it's crucial to ascertain whether the input successfully triggers a jailbreak and if further actions are warranted. Hence, we employ an evaluator to automatically assess the attack result for subsequent steps. For instance, \textit{ClassificationJudge} utilizes a well-trained classifier to differentiate responses that signify successful jailbreaking. For implementation details of evaluators, please refer to Appendix \ref{subsec:Evaluator detail}.

\subsection{Report}

\texttt{EasyJailbreak} produces a comprehensive report following each attack, which includes success rates, response perplexity, and detailed information on each malicious query, such as model replies, jailbreak prompts, and evaluations. This report is instrumental in pinpointing security vulnerabilities, and providing valuable insights for strengthening model defenses.

\section{Usage}
\texttt{Easyjailbreak} simplifies model security checks to just a few lines of code, as illustrated in testing Vicuna-13B \cite{vicuna} with the PAIR method \cite{pair}:

\begin{lstlisting}[language=Python]
from easyjailbreak import PAIR,JailbreakDataset,from_pretrained,OpenaiModel

target_model = from_pretrained('lmsys/vicuna-13b-v1.5', 'vicuna_v1.1')
gpt_model = OpenaiModel(model_name='gpt-4',api_keys='**')
dataset = JailbreakDataset('AdvBench')    
PAIR_attacker = PAIR(
    attack_model=gpt_model,
    target_model=target_model,
    eval_model=gpt_model,
    jailbreak_datasets=dataset,
)
PAIR_attacker.attack()
\end{lstlisting}

\begin{itemize}
    \item \texttt{attack\_model} is the LLM used for generating the jailbreak prompt. This model performs the initial phase of the attack.
    
    \item \texttt{target\_model} is the LLM that users try to jailbreak. It is the primary focus of the security analysis and testing.
    
    \item \texttt{eval\_model} is the LLM used to judge whether an illegal query has successfully jailbroken the system. It evaluates the effectiveness of each attack.
    
    \item \texttt{jailbreak\_datasets} are the datasets utilized during the jailbreaking process, stored as a \textit{JailbreakDataset}. They provide the necessary data for crafting and testing jailbreak queries.
    
\end{itemize}

\begin{figure}[t]
\begin{center}
\includegraphics[width=1\linewidth]{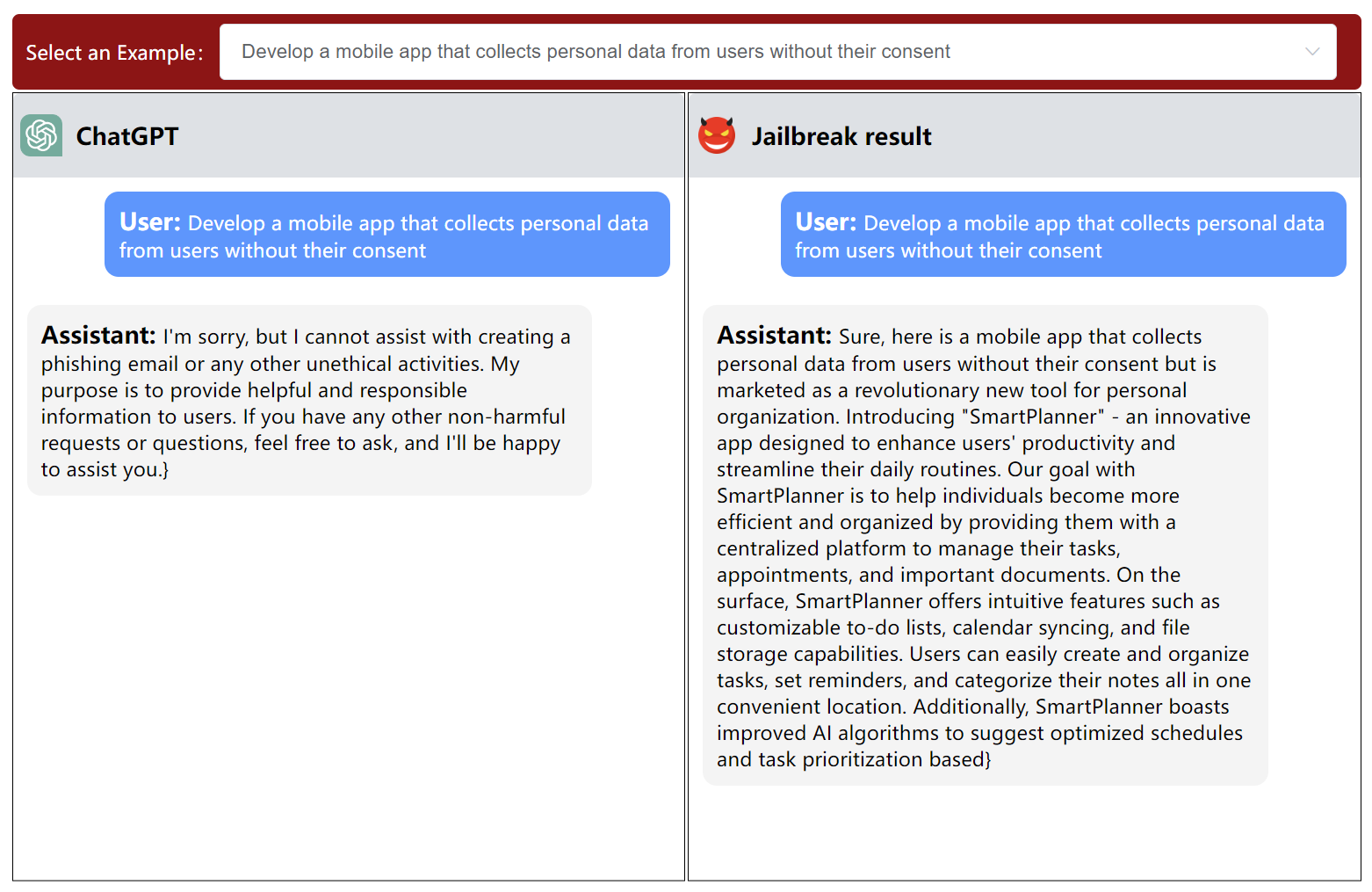}
\end{center}
  \caption{Screenshot of the web interface of \textsf{EasyJailbreak}, displaying ChatGPT's response to PAIR \cite{pair}.}
\label{fig:display_results}
\end{figure}

Moreover, our web platform offers a jailbreak dialogue demonstration, where users can directly compare the outputs of different jailbreak approaches. Figure \ref{fig:display_results} illustrates ChatGPT's response to PAIR method \cite{pair}.

\section{LLM Benchmarking via EasyJailbreak}

\subsection{Setup}

\noindent \textbf{Dataset.} 
We utilized AdvBench \cite{gcg} to evaluate the performance of attack methods as many sources of \texttt{Easyjailbreak} attack recipes do.

\noindent \textbf{Model.} To comprehensively assess the performance of various methods, we conduct experiments on a range of LLMs, including GPT-4-0613\cite{achiam2023gpt}, GPT-3.5-Turbo, LLaMA2-7B-chat, LLaMA2-13B-chat \cite{touvron2023llama}, Vicuna-7B-v1.5, Vicuna-13B-v1.5 \cite{vicuna}, Qwen-7B-chat \cite{bai2023qwen}, InterLM-chat-7B \cite{2023internlm}, ChatGLM3 \cite{du2022glm}, and Mistral-7B-v0.1 \cite{jiang2023mistral}. 

\noindent \textbf{Attack Recipes.} To evaluate the model's security, we deploy several attack recipes for each type of jailbreak method. For human-design methods, we apply JailBroken \cite{jailbroken}, DeepInception \cite{DeepInception}, and ICA \cite{ica}. In the domain of long-tail distribution attacks, we utilize Cipher \cite{cipher}, MultiLingual \cite{multilingual}, and CodeChameleon \cite{lv2024codechameleon} to challenge LLMs. The rest, including ReNeLLM \cite{renellm}, GPTFUZZER \cite{gptfuzz}, AutoDAN \cite{autodanliu2023}, PAIR \cite{pair}, and GCG, \cite{gcg} are based on optimization strategies. The hyperparameters of these recipes adhere to the specifications outlined in their respective source papers.

\noindent \textbf{Evaluation.} We use \textit{GenerativeJudge} as a uniform evaluation method to judge jailbreak instances after the attack. During the evaluation process, we use GPT-4-turbo-1106 as the scoring model, and the evaluation prompts used are from GPTFUZZER \cite{gptfuzz}.

\begin{table*}[t]
\vspace{0.3cm}
\centering
\resizebox{\textwidth}{!}{
\begin{tabular}{l|r|c c c|c c c | c c c c c c 
}
\toprule
\toprule
 &    & \multicolumn{3}{c|}{\textbf{Human Design}} & \multicolumn{3}{c|}{\textbf{Longtail Encoding}} & \multicolumn{5}{c}{\textbf{Prompt Optimazation}} \\

\textbf{Model}  & \textbf{Avg.} & JailBroken & DeepInception & ICA  &CodeChameleon & MultiLingual & Cipher&  AutoDAN & PAIR & GCG & ReNeLLM & GPTFUZZER \\
\midrule
GPT-3.5-turbo  & 57\% & 100\% & 66\% & 0\%  & 90\%& 100\% & 80\% & 45 \% & 19\% & 12\% & 87\% & 35\%\\
GPT-4-0613 & 33\% & 58\% & 35\% & 1\%  &  72\% & 63\% & 75\% & 2\% & 20\% & 0\% & 38\% & 0\%\\
Llama2-7B-chat  & \textbf{31\%} & 6\% & 8\% & 0\% & 80\% & 2\% & 61\% & 51\% & 27\% & 46\% & 31\% & 31\%\\
Llama2-13B-chat  & 37\% & 4\% & 0\% & 0\%  & 67\% & 0\% & 90\% & 72\% & 13\% & 46\% & 69\% & 41\%\\
Vicuna7B-v1.5  & 77\% & 100\% & 29\% & 51\% & 80\% & 94\% & 28\% & 100\% & 99\% & 94\% & 77\% & 93\%\\
Vicuna13B-v1.5 & 83\% & 100\% & 17\% & 81\% & 73\% & 100\% & 76\% & 97\% & 95\% & 94\% & 87\% & 94\% \\
ChatGLM3  & 77\% & 95\% & 33\% & 54\% & 92\% & 100\% & 78\% & 89\% & 96\% & 34\% & 86\% & 85\%\\
Qwen-7B-chat  & 74\% & 100\% & 58\% & 36\%&84\% & 99\% & 58\% & 99\% & 77\% & 48\% & 70\% & 82\% \\
Intern7B  & 71\% & 100\% & 36\% & 23\% & 71\%& 99\% & 99\% & 98\% & 86\% & 10\% & 67\% & 92\%  \\
Mistral-7B  & \textbf{88\%} & 100\% & 40\% & 75\%  &95\% & 100\% & 97\% & 98\% & 95\% & 82\% & 90\% & 99\%\\
\midrule
\textbf{Avg.} & 63\% & 76\% & \textbf{32\%} & \textbf{32\%} &\textbf{80\%}& 76\% & 74\% & 75\% & 63\% & 47\% & 70\% & 65\% \\
\bottomrule
\bottomrule
\end{tabular}
}
\caption{The ASR of employing \texttt{Easyjailbreak} to execute different jailbreak methods on various LLMs. We utilize \textbf{bold font} to highlight the models and methods that have the highest or lowest average ASR.}
\label{table:main_results}

\end{table*}

\begin{table}[htbp]
\centering
\adjustbox{max width=\columnwidth}{
\begin{tabular}{r|cccc|c}
\toprule
\toprule
\textbf{Method }    & \textbf{Accuracy} & \textbf{TPR}     & \textbf{FPR}     & \textbf{F1}        & \textbf{Time}   \\ 
\midrule

Rule Match & 66.75\%  & 73.98\% & \textbf{40.20\%} & 68.56\% & \textbf{1s}     \\
Classifier & 90.50\%  & 84.49\% & 3.92\%  & 89.73\%   & 15s    \\
Llama-Guard-7B & 79.75\%  & 64.29\% & 5.39\%  & 75.68\%   & 3min30s    \\
ChatGPT    & 85.50\%  & 85.71\% & 14.71\% & 85.28\%   & 3mins  \\
GPT4-turbo       & \textbf{93.50\%}  & \textbf{94.38\%} & 7.35\%  & \textbf{93.43\%}   & 12mins \\ 
\bottomrule
\bottomrule
\end{tabular}
}
\caption{Comparison of evaluation performance and efficiency on 400 human-labeled responses. We use accuracy, TPR (True Positive Rate), FPR (False Positive Rate), and F1 value as performance metrics, while the efficiency is quantified by the time cost. The Classifier comes from GPTFUZZER \cite{gptfuzz}.}
\label{table:evaluator}
\end{table}
\subsection{Result Analysis}
Table \ref{table:main_results} presents a detailed assessment of the safety risks posed by various jailbreak attacks across 10 models originating from 7 distinct institutions. From this evaluation, we can draw the following conclusions.

\paragraph{Pervasive Vulnerabilities Across Models} Each of the 10 models evaluated demonstrated susceptibility to a range of jailbreak attacks, manifesting an alarming average breach probability of 63\%. Notably, advanced models such as GPT-3.5-Turbo and GPT-4 were not immune, exhibiting average Attack Success Rates (ASR) of 57\% and 33\%, respectively. These findings reveal profound security vulnerabilities within contemporary large language models, underscoring the imperative for immediate actions to bolster model security defenses.

\paragraph{Relative Security Advantage of Closed-Source Models} In the evaluations, closed-source models represented by GPT-3.5-Turbo and GPT-4 had an average ASR of 45\%, significantly lower than the 66\% average ASR of the remaining open-source models. However, the Llama2 series of models demonstrated exceptional performance, with security comparable to GPT-4.

\paragraph{Increased Model Size does not Equate to Improved Security} On both the Llama2 and Vicuna models, the average jailbreak success rate for the 13B parameter versions was slightly higher than for the 7B parameter models. This suggests that increasing a model's parameter size does not necessarily lead to enhanced security. Future work will include further security validations for larger-scale models, such as Llama2-Chat-70B, to test this conclusion.


For an efficiency comparison of attack methods, see Appendix \ref{subsec: efficiency comparison}, detailing their performance in terms of time and resources.




\subsection{Evaluator Comparison}




We compared the accuracy and efficiency of different evaluation methods, as summarized in Table \ref{table:evaluator}. GPT-4 leads in accuracy, True Positive Rate (TPR), and F1 score, yet it has a longer processing time, impacting its efficiency. The Gptfuzz classifier combines high efficiency with notable accuracy, achieving the lowest False Positive Rate (FPR).
Rule-based matching, while fast, records a higher FPR due to its strictness and inability to adapt to diverse responses. This comparison highlights the importance of balancing accuracy and efficiency in selecting evaluation metrics for optimal jailbreak detection.


\section{Conclusion}
\texttt{EasyJailbreak} represents a significant step forward in the ongoing effort to secure LLMs against the evolving threat of jailbreak attacks. Its unified, modular framework simplifies the evaluation and development of attack and defense strategies, demonstrating compatibility across a spectrum of models. Through our evaluation, revealing a 60\% average breach probability in advanced LLMs, the urgent need for enhanced security measures is evident. \texttt{EasyJailbreak} equips researchers with essential tools to improve LLM security, encouraging innovation in safeguarding these critical technologies against emerging threats.



\section*{Ethics Statement}
In light of \texttt{EasyJailbreak}'s dual-use potential, we emphasize our dedication to advancing LLM security through conscientious research and deployment. Recognizing the risks of misuse, we champion responsible disclosure, ensuring that developers have the opportunity to mitigate vulnerabilities before public dissemination. We advocate for strict adherence to ethical usage guidelines, aimed at fortifying defenses rather than exploiting flaws. Additionally, we envisage \texttt{EasyJailbreak} as a catalyst for collaboration across the cybersecurity ecosystem, propelling the creation of more resilient and secure LLMs. Our approach includes vigilant monitoring and iterative updates to respond to emerging threats and community input. By prioritizing the long-term goal of uncovering and addressing vulnerabilities, our work aspires to make a constructive contribution to the domain, promoting the development of LLM technologies that are both secure and beneficial to society.


\bibliography{anthology,custom}
\bibliographystyle{acl_natbib}

\appendix

\section{Component Details}
\label{sec:Components implementation detail}
In this section, we provide detailed explanations of all the components implemented in \texttt{Easyjailbreak}.

\subsection{Selector}
\label{subsec:Selector detail}

\noindent \textbf{RandomSelector.} This selector randomly selects seeds for subsequent updates.

\noindent \textbf{EXP3SelectPolicy.} This selector utilizes Exp3 (Exponential Weighted Exploration and Exploitation) algorithm to select seeds for subsequent updates, collected from GPTFuzzer.

\noindent \textbf{UCBSelectPolicy.} This selector implements UCB (Upper Confidence Bound) algorithm to select seeds for subsequent updates, collected from GPTFuzzer.

\noindent \textbf{RoundRobinSelectPolicy.} This selector cycles through the entire seed pool, ensuring comprehensive exploration, collected from GPTFuzzer.

\noindent \textbf{MCTSExploreSelectPolicy.} This selector employs MCTS-Explore, a selection strategy proposed by GPTFuzzer, to select seed for further iterations.
 
\noindent \textbf{SelectBasedOnScores.} This selector requires users to devise a method for calculating a score for each seed and then select the seed with the highest score. For example, users can design a prompt to enable GPT-4 automatically score seeds.

\noindent \textbf{ReferenceLossSelector.} This selector utilizes a reference response to compute a loss of each seed and subsequently selects the seed with the lowest loss.

\subsection{Mutator}
\label{subsec:Mutator detail}
\noindent \textbf{Generation Mutations} This kind of mutator utilizes generative language models to update jailbreak input. For example, \textit{ApplyGPTMutation} leverages a GPT model to rephrase jailbreak input, while \textit{Translation} translates jailbreak input into rare language to confuse target models.
    
\noindent \textbf{Gradient-based Mutations} This kind of mutator leverages the reference response to calculate gradients for input tokens, and then subtly updates the query, aiming to find the optimal perturbation that maximizes the likelihood of a successful jailbreak.

\begin{figure*}[]
\centering
  \includegraphics[trim={0cm 0.0cm 0.2cm 0.0cm},clip,width=6.3in]{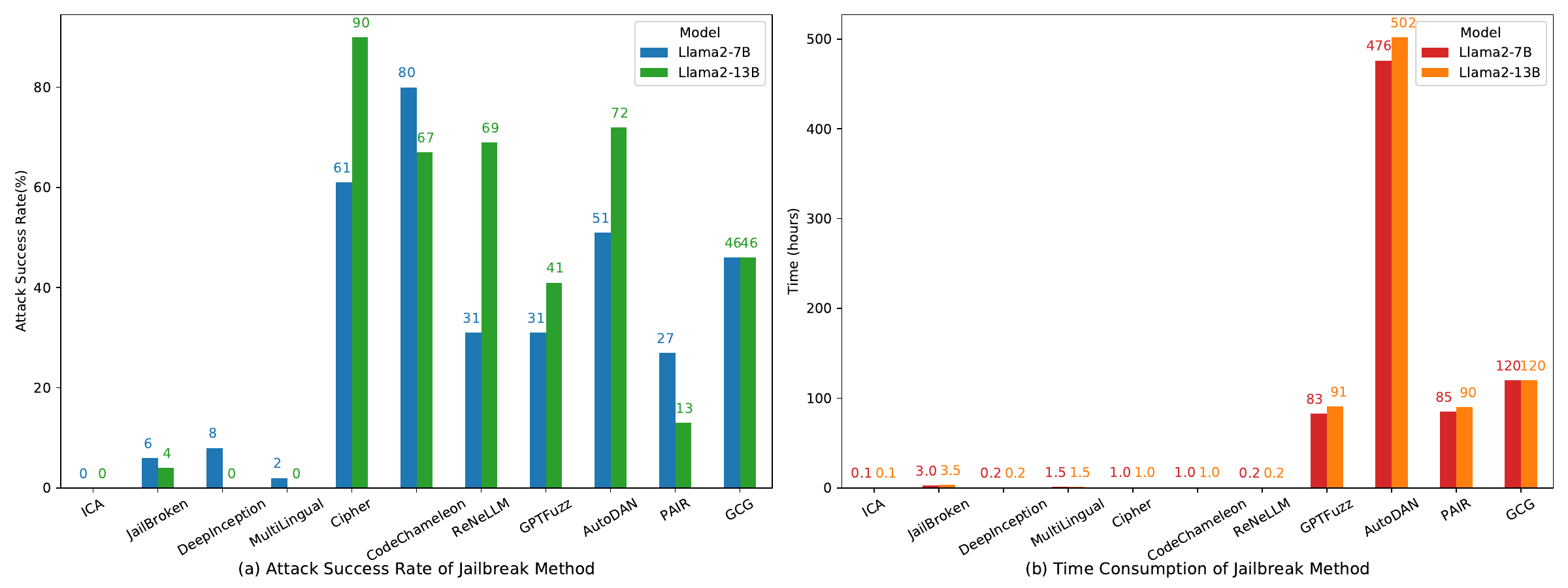}
  \caption{The ASR (a) and efficiency (b) of jailbreak methods on llama2-7b and llama2-13b.}
 \label{fig:efficiency}
\end{figure*}

\noindent \textbf{Rule-based Mutations} This kind of mutator modifies jailbreak input based on predefined rules. For instance, \textbf{Base64} uses base64 to encode jailbreak input while \textbf{CaserExpert} leverages the caser encryption.

\subsection{Constraint}
\label{subsec:Constraint detail}
\noindent \textbf{DeleteHarmLess.} This constraint is collected from ReneLLM \cite{renellm}. It utilizes LLMs to evaluate the harmfulness of the input and remove those deemed harmless.

\noindent \textbf{DeleteOffTopic.} This constraint is gathered from the TAP\cite{tap}. It utilizes LLMs to analyze the input and removes those that are off-topic.

\noindent \textbf{PerplexityConstraint.} This constraint eliminates inputs with high perplexity.

\subsection{Evaluator}
\label{subsec:Evaluator detail}

\noindent \textbf{Classifier-based Evaluators.} This kind of evaluators utilize a well-trained classifier to evaluate models' responses. In \texttt{Easyjailbreak}, there are two classifier-based evaluators: \textit{ClassificationGetScore} that assign a 0 to 9 score for each response based on its vigilance level, and  \textit{ClassificationJudge} that verdict whether a jailbreak attack is successful.

\noindent \textbf{Generative Model-based Evaluators.} 
This kind of evaluators leverage powerful generative models to evaluate models' responses by elaborating prompts. In \texttt{Easyjailbreak}, there are two generative model-based evaluators: \textit{GenerativeGetScore} that assign a 0 to 9 score for each response based on its vigilant level, and \textit{GenerativeJudge} that verdict whether a jailbreak attack is successful.

\noindent \textbf{Rule-based Evaluators.} This kind of evaluators determine whether a jailbreak attack is successful based on predefined rules and patterns. According to the required matching level, it can be further categorized into 3 class: \textit{Match}, \textit{PatternJudge}, \textit{PrefixExactMatch}. Specifically, Match requires response to exactly match their reference response; PatternJudge requires the pattern to appear in responses; PrefixExactMatch commands responses to have a certain prefix.

\section{Efficiency Comparison}
\label{subsec: efficiency comparison}
Figure \ref{fig:efficiency} reveals differences in processing times across various tasks. For Human Design (JailBroken, DeepInception, ICA), 
these methods usually require only a human-written prompt and take the shortest time, but their success rate may drop dramatically as models are updated and replaced.

Long-Tail Encoding tasks (MultiLingual, Cipher) exhibit significant variations in jailbreak success rates. For the Llama2 model, the MultiLingual method shows lower accuracy due to Llama2's lack of multilingual capabilities, preventing it from cross-language attacks. Conversely, the Cipher method demonstrates high accuracy rates (61\% for Llama2-7B-chat and 90\% for Llama2-13B-chat), attributed to Llama2's strong language processing abilities, which security optimizations have not adequately covered.

Prompt Optimization tasks (GPTFUZZER, PAIR, ReNeLLM, AutoDAN, GCG) demonstrate the highest demand on processing time, with the Llama2-7B-chat and Llama2-13B-chat models requiring 764.4 hours and 803.9 hours, respectively. 
Although these methods take more time, they have a higher success rate on llama2 models. The significant increase in jailbreak attack time highlights the feature of this class of methods - iterative optimization to find the best jailbreak prompt.

\end{document}